\documentclass[12pt, conference]{IEEEtran}
\ifCLASSINFOpdf
\else
\fi
\usepackage{graphicx}
\usepackage{amsmath}
\usepackage{booktabs}

\begin{document}
%
\title{Structure-guided Image Outpainting}

\author{\IEEEauthorblockN{Xi Wang}
\IEEEauthorblockA{\small University of Alberta\\
Computing Science - Multimedia\\
Student ID: 1773819\\
Email: xi29@ualberta.ca}
\and
\IEEEauthorblockN{Weixi Cheng}
\IEEEauthorblockA{\small University of Alberta\\
Computing Science - Multimedia\\
Student ID: 1595976\\
Email: weixi@ualberta.ca}
\and
\IEEEauthorblockN{Wenliang Jia}
\IEEEauthorblockA{\small University of Alberta\\
Computing Science - Multimedia\\
Student ID: 1506232\\
Email: wenliang@ualberta.ca}}


%


\maketitle

\begin{abstract}
Deep learning techniques have made considerable progress in image inpainting, restoration, and reconstruction in the last few years. Image outpainting, also known as image extrapolation, lacks attention and practical approaches to be fulfilled, owing to difficulties caused by  large-scale area loss and less legitimate neighboring information. These difficulties have made outpainted images handled by most of the existing models unrealistic to human eyes and spatially inconsistent. When upsampling through deconvolution to generate fake content, the naive generation methods may lead to results lacking high-frequency details and structural authenticity. Therefore, as our novelties to handle image outpainting problems, we introduce structural prior as a condition to optimize the generation quality and a new semantic embedding term to enhance perceptual sanity. we propose a deep learning method based on Generative Adversarial Network (GAN) and condition edges as structural prior in order to assist the generation. We use a multi-phase adversarial training scheme that comprises edge inference training, contents inpainting training, and joint training. The newly added semantic embedding loss is proved effective in practice.
\end{abstract}


\begin{figure*}[t!]
\centerline{\includegraphics[width=15cm]{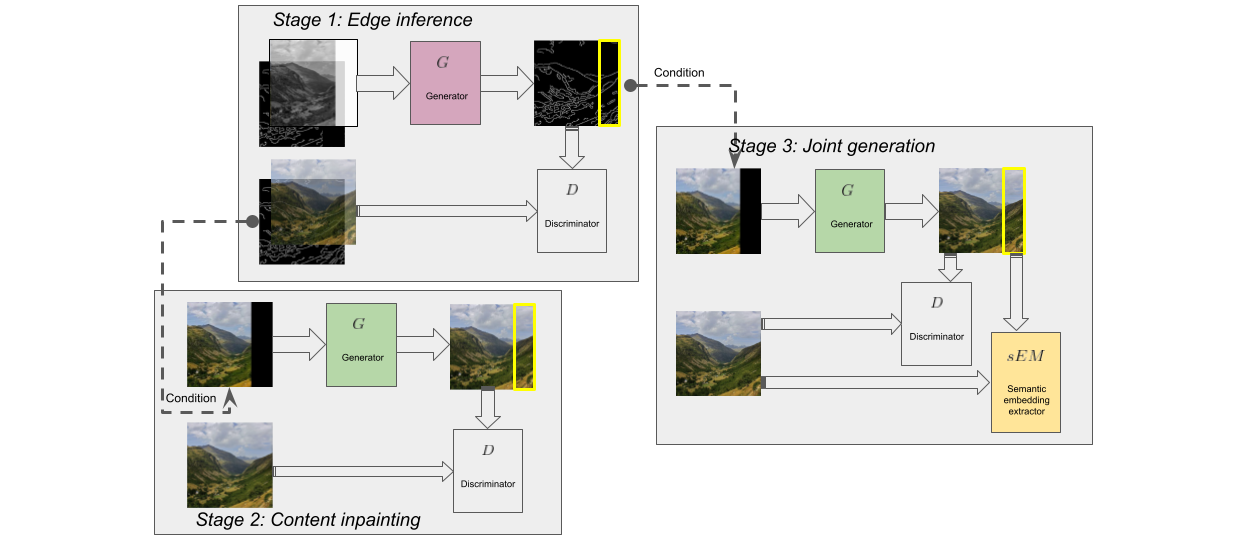}}
\caption{General pipeline.}
\label{general}
\end{figure*}

%
\IEEEpeerreviewmaketitle

\section{Introduction}
With the development of deep learning techniques, it has led to a number of new applications in the field of computer vision. Image processing plays a very important role in these various applications, such as image inpainting, image restoration and image reconstruction. Image inpainting aims to improve or restore the quality of parts of an image and requires the completion of missing areas of the image while preserving the image context. However, compared to image inpainting, which is used to fill in missing regions in an image, image outpainting is a considerably more difficult challenge. The problem explored in this paper is how to implement image outpainting on the basis of deep learning techniques.\par Image outpainting is the process of extending the exterior of a given image to preserve the background of the original image by considering the existing visual elements of the image, including shadows, reflections, and textures. The reason for the difficulty of this task is due to large-scale area loss and less legitimate neighboring information in the image outpainting scene. Moreover, most existing models process images that lack not only visual quality but also structural consistency. \par In this research, a novel approach is introduced in order to solve the above problems, make the outpainted images more realistic for the human eye and provide better structural information in advance. We propose a deep learning method based on generative adversarial networks (GAN) by using condition edges as structural prior. We use a three-stage adversarial training scheme, where the structure-guided network is implemented using edges predicted by a trained edge inference model as conditions to guide generation. And in order to optimize the high-frequency details of the generated results, improve the authenticity of the structure, and preserve the sanity of perception, we introduce a new loss function term. The holist system architecture is shown as Fig.~\ref{general}. Based on our evaluation results, we observe that by introducing the new loss function, both PSNR and SSIM significantly improve and MAE decreases.\par We implement image outpainting on the landscape dataset because it is of strong practical importance to extrapolate landscape images. For example, in-camera cropping by the camera for anti-shake can cause the captured image to be zommed in and missing the desired details. And the captured image can be directly expanded when the content framing range has limitations.

\vspace{4mm}
\section{Literature review}
\subsection{Multimedia-related applications}
\subsubsection{RCA-NET: Image Recovery Network with Channel Attention Group for Image Dehazing}
To provide more visible information features and more significant details to the restored image, Du et al.~\cite{du2019rca} proposed an end-to-end dehazing network that uses a channel attention model to recover fog-free images from fogged images and is supervised using perceptual loss. The channel attention model is given normal convolution weights for further feature extraction. The atmospheric scattering model in RCA-Net is simplified during the image restoration process by providing the channel attention model with normal convolution weights. This simplification of the haze removal network allows the channel attention model to focus on more important information for further feature extraction as a way to retain more realistic color and structural details in the image. 
\subsubsection{Traffic Image Dehazing Based on Wavelength Related Physical Imaging Model}
Considering the effect of the concentration of fog on image dehazing, Wang et al.~\cite{wang2019traffic} proposed a dehazing method based on the wavelength-related physical imaging model. A transmission estimation strategy is designed based on the setting that the colors of objects in the image are determined according to the reflection of different wavelengths as a priori. A transmission map containing continuous scene information can be obtained using this method by segmenting data using the maximum fuzzy correlation and the graph cut algorithm, which is then used as a guide image in a guide filter. This method avoids the misclassification of scene items since it uses both threshold-based and space-based segmentation techniques. Additionally, a proposed iterative approach is made to enhance fuzzy correlation segmentation's computational effectiveness. 
\subsubsection{ProDeblurGAN: Progressive Growing of GANs for Blind Motion Deblurring in Face Recognition}
As the progressive growth of GANs progresses, Mahalingaiah and Matichuk~\cite{mahalingaiah2019prodeblurgan} took advantage of this progressive, adversarial training to propose a progressive growth motion deblurring network, ProDeblurGAN, designed explicitly for the motion deblurring of single-face photographs. This model generates higher-quality deblurred images. Rather than a one-time initialization, the experiments gradually increased the number of layers during training, generating high-resolution images. The accuracy of face recognition is improved by increasing the quality of the captured images so that finer facial details, and potential information, are considered features and used in the learning process. In addition, instead of taking the random noise found in traditional GANs, the proposed generator will take the blurred images as input.
\subsubsection{Semantic Learning for Image Compression (SLIC)}
Mahalingaiah et al.~\cite{mahalingaiah2019semantic} proposed a deep learning convolutional neural network, which was implemented and trained for image recognition and image processing tasks. To improve the visual quality in lossy compression, the semantic graph of an image is generated tailor-made by slightly increasing the complexity of the encoder to make the encoder content-aware for a specific image. The presented work proposes a novel architecture explicitly designed for image compression, which builds semantic graphs for salient regions, allowing them to be encoded with higher quality than background regions.
\subsubsection{Synthetic Aerial Image Generation and Runway Segmentation}
Sharma et al.~\cite{sharma2019synthetic} proposed a system to generate 2D images of the runway taken during takeoff and landing and to synthesize the airport using a 3D model to enhance the visual perception of pilots. The 3D features of the vanishing point and the runway start will be efficiently communicated to the pilot through the synthetic image sequences obtained from the 3D model of the view. A dataset of image sequences for the airport and runway is provided by first creating an accurate 2D representation of the flight sequence and rendering it in a virtual environment using a 3D model. The takeoff and landing routines are then projected as two-dimensional visuals after the flight sequences are simulated using the aircraft's GPS and AHRS data.
\subsubsection{Fog Removal of Aerial Image Based on Gamma Correction and Guided Filtering}
Liu et al.~\cite{liu2019fog} proposed an aided vision system combining Gamma correction and Retinex defogging algorithm, which delivers a clearer, higher contrast, and consistent color image of foggy aerial skies through processing. In the experiments based on the SSR model, the original images are corrected using Gamma as the guide map to correct the contrast and brightness of the original images. Then the images with various Gamma corrections are then submitted to tri-scale bootstrap filtering. The fused filtered images are fed into the single-scale Retinex model to obtain the pre-set aerial defogging images.
\subsubsection{Hough Transform for Feature Detection in Panoramic Images}
Most mirror systems used in the camera are not meet the Single Viewpoint (SVP) criteria. It will result in detecting straight lines becoming difficult by using a pinhole camera model. To facilitate this problem and detect straight-line features in panoramic non-SVP images, Fiala and Basu~\cite{fiala2002hough} proposed an approach using the Panoramic Hough transform, designed for catadioptric panoramic sensors with spherical mirrors. The experiment results showed that this is a way to identify horizontal lines to supplement the trivial detection of vertical lines. It also demonstrated the robust performance of this transform in detecting straight-line features with estimated calibration.
\subsubsection{QoE-Based Multi-Exposure Fusion in Hierarchical Multivariate Gaussian CRF}
Shen at al.~\cite{shen2012qoe} proposed a fusion algorithm based on the hierarchical multivariate Gaussian conditional random field model to show improved performance for multi-exposure fusion. Perceived local contrast and color saturation are two perceptual quality measures used in this model. Specifically, this model calculates the contrast in the luminance channel of the LHS color space to preserve details. To get satisfactory results for color images, this model also uses color saturation to measure the colorfulness of a given pixel and denote red, green, and blue components in the RGB color space. The experiment results showed that the proposed algorithm could provide viewers better quality of experience in fused images.
\subsubsection{Panoramic Video with Predictive Windows for Telepresence Applications}
Baldwin et al.~\cite{baldwin1999panoramic} applied a predictive Kalman filter to display panoramic images in an effective telepresence system. This telepresence system integrates a panoramic imaging system and the viewing direction, which the Kalman filter predicts. In this telepresence system research, the authors present the experiment results from the prediction process, as well as the experiment results from operator experience. The authors pointed out that appropriate error modeling and parameter identification can improve the prediction process and give better quantitative results. The local image information will be used to simulate continuously flowing pictures when transmitting images from a remote site with delays.
\subsubsection{XAI Feature Detector for Ultrasound Feature Matching}
Based on the Deep Unfolding Super-Resolution Network (USRNET), Wang et al.~\cite{wang2021xai} proposed an algorithm using Explainable Artificial Intelligence (XAI) to detect features in ultrasound images. This algorithm consists of two steps: feature detection and feature matching. Specifically, the first step implements the XAI feature detector using USRNET to generate gradients and guided back-propagation to generate feature maps based on these gradients' information. Unlike the original application of guided back-propagation with classification networks, the authors improved it to reveal the features that significantly impact the super-resolution network. According to these detected features, the second step implements the feature matching method through the Robust Independent Elementary Features (BRIEF) descriptor and Brute-force feature matcher. The authors pointed out that because of the nature of USRNET, this algorithm needs to be improved to detect entire images rather than the high-frequency areas only.  

\subsubsection{Nose Shape Estimation and Tracking for Model-Based Coding}
Yin and ~\cite{yin2001nose} proposed three steps to implement a feature detection method on the facial organ areas. The first step is to limit the facial feature detection region to certain areas, and a two-stage region-growing algorithm is used to make the detection less sensitive to noise. A significant global growing threshold will be used to explore more skin areas in the first stage. In contrast, the second stage will estimate region information in the individual feature areas. The pre-defined templates are used in the second step to extract the shape of the nostril and nose side. After detecting the facial features and extracting the nose’s shape, a 3D wireframe model will be used to track the motion of facial expressions. The experiment results demonstrated that the proposed method could detect most nose features correctly.
\subsubsection{Dynamic Deep Pixel Distribution Learning for Background Subtraction}
To improve the background subtraction method, Zhao and Basu~\cite{zhao2019dynamic} proposed a Dynamic Deep Pixel Distribution Learning (D-DPDL) model based on automatically learning the distribution. This model consists of a Random Permutation of Temporal Pixels (RPoTP) distribution descriptor and a Bayesian refinement model. Specifically, the RPoTP distribution descriptor is used to force the convolutional neural network to learn statistical distribution. Since the random noise could be generated during the random permutation process, a Bayesian refinement model will be used to handle noise. Limited ground-truth frames can be used to train the proposed model to capture RPoTP features. The experiment results showed that the proposed model could also be used to train and test videos because of the nature of the statistical distribution.

\subsection{Generative methods}
Machine learning can be broadly classified into two types in terms of learning approaches: supervised learning and unsupervised learning. For tasks such as classification, training on a labeled dataset makes it possible to construct a mapping of data to labels, which is the goal of supervised learning. For unsupervised training, the purpose of training is to find the intrinsic features or probability distribution of the data, which is suitable for applications such as clustering, feature learning, or dimensionality reduction. Unsupervised learning can be further divided into non-probabilistic models and probabilistic models. The predecessor of the popular variational auto-encoder (VAE)~\cite{kingma2013auto} is the auto-encoder~\cite{rumelhart1985learning} from the non-probabilistic model approach. An auto-encoder consists of an encoder and a decoder, where the encoder first compresses the input into a low-dimensional latent vector. Then a decoder composed of several deconvolution layers recovers the hidden vector into a high-dimensional output. The goal of the auto-encoder is to minimize the distance between the output and the input so that it can be used for reconstruction. Still, due to the lack of regularity of the latent vector, it is impossible to generate new data through the manipulation of the latent vector. The VAE replaces the latent representations with probability distributions and then reconstructs the representation by sampling it over the latent probability distribution to obtain the reconstruction loss. The probability distribution allows the VAE to artificially control the model output for the purpose of generating new samples. In addition, the mapping relationship and optimization process of VAE are explicitly modeled by a perfectly provable mathematical model, but this creates a strong preset in the design of VAE, and its optimization process compulsorily fits the data to a finite-dimensional mixed Gaussian distribution, leading to inevitable information loss and blurring of the generated picture. There is another generative model; unlike that VAE has intractable likelihoods, PixelRNN/CNN~\cite{van2016pixel}~\cite{van2016conditional} is tractable. The PixelRNN/CNN computes the maximum likelihood of the current pixel from top to bottom from left to right, so each pixel can be traced back to its previous context, which is a sequential computation.
The objective function of PixelRNN/CNN is to find the exact maximum likelihood by the chain rule, which makes the training of PixelRNN/CNN stable, but the considerable computation also renders the fitting and inference process very slow though. Both of the above-mentioned generative models are explicit, except for the generative adversarial network (GAN), which is implicit. The GAN is inspired by the zero-sum game in game theory. It consists of a generator $G$ and a discriminator $D$. The generator generates samples starting from a noise sampled from a Gaussian distribution, which is further fed into the discriminator for distinguishing. The discriminator will determine only whether the image generated by the generator is true or not by a probability that takes a value between 0 and 1, thus obtaining a loss. In other words, $G$ captures the actual distribution of the samples, while the $D$ is a binary classifier that determines whether the input is real data or a generated sample.
The objective function of GAN is to minimize the $G$ loss and maximize the $D$ loss in order to ultimately make the probability distribution of the generated samples as close as possible to the real data probability distribution. GAN, which is implicit, does not have an obvious mapping relationship, making it not very interpretable. The structure of the $G$ and $D$ proposed by Ian in 2014~\cite{goodfellow2014generative} is realized by a Multi-layer Perceptron (MLP) with a multi-layer fully connected network as the main body. However, it has problems of difficulties in adjusting the parameters, training failures, and low quality of the generated images, especially for complex datasets. Since the Convolutional neural network (CNN)~\cite{krizhevsky2012imagenet} has a more robust fitting and representation capabilities than MLP and has achieved outstanding results in discriminative models, Alec et al.~\cite{radford2015unsupervised} introduced CNNs into generators, and discriminators called Deep Convolutional Adversarial Neural Network (DCGAN). In addition, the importance and guiding role of hidden layer analysis and visual counting for GAN training is also emphasized by Alec. Although DCGAN does not optimize the interpretability of GAN, its powerful image generation effect attracts more researchers to focus on GAN, proves its feasibility and provides experience, and provides a reference to later researchers for neural network structure.
\subsubsection{Conditional GAN}
Mirza and Osindero proposed conditional GAN~\cite{mirza2014conditional} in 2014, which opened up a broader world for GAN-based research and applications. The original GAN does not have any conditional constraints, and the generated images are random. Therefore, the authors considered adding some conditional information, such as category labels or other types of data, which are encoded and fed into the network together with the original noise $z$ so that image generation can proceed in a prescribed direction. Conditional generation usually has not only a potential distribution but also a determined conditional variable for inputting to the generator, achieving such as the text-to-image task or the semantics-to-image task. Conditional GANs can be used for tasks such as style transferring, directed generation, etc. EdgeConnect, the generative network adopted in this thesis, is a conditional GAN that generates content more in line with topology and local details by conditioning on a low-abstraction and low-complexity edge prior.
\subsection{Image inpainting}
Pathak et al.~\cite{pathak2016context} propose an approach that is the pioneer in the image inpainting field to deploy deep learning. An encoder-decoder architecture is proposed to implement an unsupervised learning method based on contextual pixel prediction. The encoder-decoder architecture is coupled with convolutional neural networks and GAN to generate images with its semantic information. The encoder is derived from the AlexNet architecture, and a Channel-wise fully-connected layer is proposed and applied in the decoder instead of the fully-connected reduction parameter. This architecture also served as a prototype for various later deep learning-based implementations of image inpainting algorithms.\par
Iizuka et al.~\cite{iizuka2017globally} proposed a method to guarantee the local consistency of the generated images while having global consistency. Using a fully convolutional neural network, the authors can achieve recovery of arbitrary resolution images that fill in missing regions of any shape. To train this image inpainting network, the authors use global and local context discriminators for training to distinguish between real images and inpainting-ed images. This architecture motivates the generator to generate images that are indistinguishable from real images in terms of overall consistency and detail. While the global discriminator is responsible for the overall consistency, the local discriminator is only responsible for small regions centered on the recovered area to ensure the local consistency of the generated patch. The authenticity scores are obtained between the features encoded by the global discriminator and the features encoded by the local discriminator by concatenating them into the fully connected layer.\par
Hereto deep learning-based image inpainting algorithms have been able to generate reasonable image structures and textures. But the boundaries of generated patches often suffer distortion and blurring caused by the inability of convolutional neural networks to extract information from distant legitimate regions. Yu et al.~\cite{yu2018generative} propose a new network architecture consisting of two stages. The first stage is a dilated convolutional network backpropagated by reconstruction loss to obtain a rough restored image; the second stage uses the proposed contextual attention layer to accomplish fine inpainting. The contextual attention layer relies on the features of non-missing patches as convolutional kernels to process the generated Patches to refine the distorted and blurred primary generation.\par
Song et al.~\cite{song2018contextual} further explored on contextual-based inpainting. The method is divided into two networks as well as three steps. In the inference phase, the authors propose an Image2Feature network to fill the damaged images with rough content as inference and extract the feature maps from the primarily restored images. In the matching phase, the authors perform patch-swap on the feature maps obtained in the inference phase. Because the information of the feature maps obtained in the first stage is fuzzy in the missing regions, the authors' design diffuses the texture information from the legitimate areas to missing parts by patch-swap. In the translation phase, the authors propose a Feature2Image network to decode the feature maps back into context-based inpainting results. The holistic network is forwarding rather than an iterative optimization, which makes the method advantageous in terms of speed and computational costs.\par
The plain deep learning-based inpainting method uses the statistical information of the remaining valid part of images to fill the holes by using fixed values to replace the missing contents, which treats all pixels equally when convolving. And the convolution results are affected by the initialized values filled in the holes, causing a lack of texture in the damaged region or obvious artifacts around the holes. Liu et al.~\cite{liu2018image} propose a method for image inpainting that uses Partial Convolution to robustly handle masks of any shape, size, and location. And the performance does not deteriorate dramatically as the hole size increases. The Partial convolutions proposed in this paper make the CNN aided by a mask that distinguishes between holes and non-holes during propagation, thus improving the authenticity and fineness of the generated results.\par
However, partial convolution has some shortcomings, i.e., the mask of the partial convolution network fades away at deeper layers during the mask update process. Therefore, partial convolution can be considered as an unlearnable hard gating, for which Yu et al.~\cite{yu2019free} propose mask update criteria that can be learned automatically, i.e., gated convolution to extend partial convolution by providing a dynamic feature selection mechanism that can be learned for each channel at each spatial location of all layers. Gated convolution can be considered as learnable soft masking.\par
Another factor resulting in the poor quality of the images generated by the deep method is the lack of high frequency information aids. Nazeri et al.~\cite{nazeri2019edgeconnect, Nazeri_2019_ICCV} propose a structure-guided inpainting process to recover the missing parts of an image. This architecture first predicts the missing region's skeleton/structure (i.e., edges/contours) and then fill in the color based on the generated edges. Existing deep image inpainting methods usually produce blurred regions of low frequency, while high-frequency details may be omitted. The authors provide the generator with structural prior information of the missing region, which produces better local fine texture details. The edge map of the missing region is a good choice for a priori information because it contains the overall structural guidance of the image. The proposed architecture is divided into two networks, i.e., edge prediction and content inpainting. And the training is divided into three stages, i.e., edge prediction training, content inpainting training, and joint training. The content inpainting training uses the ground truth edge maps as prior, while the edges generated by the edge inference network trained in the first phase are used as prior in the joint training.

\subsection{Image outpainting}
As a very first attempt to deploy deep methods to image outpainting, i.e. image extrapolation, Sabini and Rusak~\cite{sabini2018painting} proposed a three-phase training schedule to stably train a Deep Convolutional Generative Adversarial Network (DCGAN) architecture. There is a loss function associated with each phase. Based on the first loss function, the generator will be trained by updating its weights in phase one. Updating the discriminator’s weight according to the second loss function will be completed in phase two. In the third phase, the generator and discriminator will be trained adversarially according to the third loss function. They also pointed out that it is necessary to use dilated convolutions to provide a sufficient receptive field to perform outpainting. Besides, a global discriminator used in training will result in a realistic result. And a local discriminator can be used to augment the network to improve the result’s quality.\par
One challenge is that image outpainting suffers meaninglessness of generated contents. To tackle with this problem, Khurana et al.~\cite{khurana2021semie} propose a new paradigm of semantic awareness to perform image extrapolation so that new object instances can be loaded into the synthesized image. The previous image outpainting approaches rely on existing entities in original images to generate new content. While this method also pays attention to adding new entities in the extended region based on the context by extrapolating the image in the semantic label space. As a similar case, Wang et al.~\cite{8953261} propose a semantic regeneration network that uses multiple spatial-related losses. This semantic regeneration network uses two sub-networks - Feature Expansion Network (FEN) and Context Prediction Network (CPN) - to extract deep features from the original image and decode these features into the resulting image along with the semantic information.\par
Cheng et al.~\cite{cheng2022inout} proposed an in and out method to implement image outpainting. Generator training and outpainting via inversion are two main stages in the in and out method. In the first stage, a generator combines the desired coordinate and the optimal latent code and generates a new patch to perform the image outpainting. Based on this trained generator, the second stage will implement an inversion process to seek multiple latent codes, recover available regions, and predict outpainting regions.\par
The idea of using some prior as a condition to guide the generation is also applied to image outpainitng tasks. Wang et al.~\cite{wang2021sketch} propose an outpainting method where the user can control the landscape image generation through a free-form sketch. The architecture consists of three modules, one for generation and two for alignment modules. The two alignment modules are used to promote the authenticity and consistency of the generated content with the provided sketches. The authors first apply a holistic alignment module to make the generated parts resemble the real parts globally. Then the sketch is generated in reverse from the generated content and the sketch alignment module is used to motivate consistency with the real sketch. This method trains a generator that will consider the user-entered sketch as a content trend-guiding prior, enhancing the detail and controllability of the generated content.

\begin{figure}[t]
\centerline{\includegraphics[width=9cm]{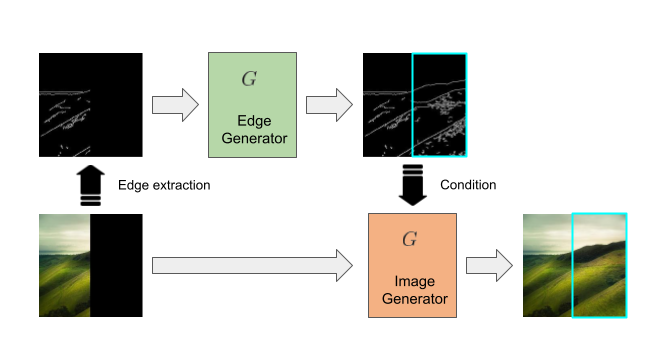}}
\caption{Our approach.}
\label{appr}
\end{figure}

\vspace{4mm}
\section{Approach}
Our overall pipeline is shown in Fig.~\ref{appr}. Details are elaborated below.

\begin{figure*}[htb!]
\centerline{\includegraphics[width=12cm]{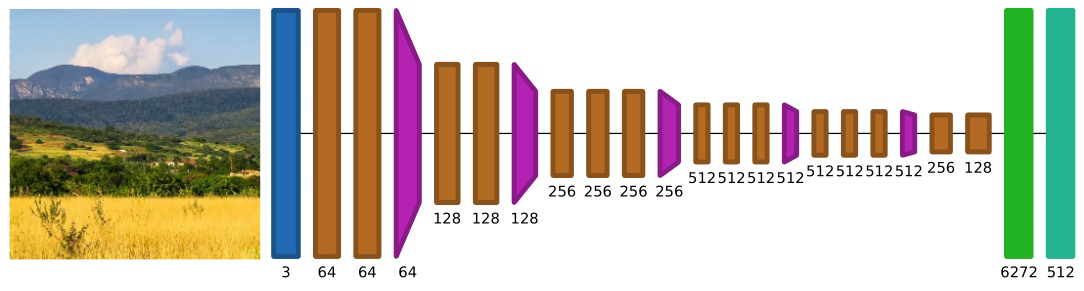}}
\caption{Network architecture of the semantic embedding extractor.}
\label{vgg16}
\end{figure*}
\subsection{Preprocessing}
We design to first generate a binary mask for occluding to-be-outpainted contents in the original images. We will multiply ground truth images with the binary mask to generate a masked dataset. Thus, a new dataset preprocessed by masking the original dataset will be used for model training. The masked dataset will be fed into the structure-guided neural network for the outpainitng of the right masked region stage by stage. \par

\subsection{Structure-guided generative network}
The generation model is transfer-learned using the state-of-art inpainting network EdgeConnect~\cite{nazeri2019edgeconnect, Nazeri_2019_ICCV}. EdgeConnect is a structure-guided network combining edge inference and content inpainting. The edge predicted can significantly help the inpainting process because the structure information is well preserved and creates a reconstructed area derived to a certain extent from the valid clues.

\subsection{Semantic embedding loss}
Preserving semantic information in images is a challenge for the picture outpainting task of landscape images. Although we have introduced structure-guided generation networks to make the generated results more visually realistic, the results generated by models using pure structure-guided networks are prone to meaningless or textures with stray colors and artifacts due to the semantic inconsistency among landscape-like pictures. To conquer this problem, we try to introduce a new semantic embedding loss term, aiming to measure the semantic discrepancy between the generated images and the ground truth by this loss.

\subsubsection{VGG16}
VGG16 is a deep neural network architecture proposed by the Visual Geometry Group, Oxford University. This network was trained on the largest image classification dataset, ImageNet. The original network output is a 1,000-dimensional fully-connected layer corresponding to the 1,000 categories of ImageNet. VGG16 consists of two blocks: the first part is a network consisting mainly of convolutional layers for feature extraction, and the second block consists of three consecutive fully-connected layers. Since we do not need to handle the classification, but only to extract the semantic features of the input images, we modified the second block of VGG16 by replacing the three fully connected layers with one convolutional layer and one fully-connected layer to reduce the number of parameters. The last fully-connected layer has a dimension of 512 and characterizes a 512-dimensional feature vector, which holds the semantic information of the input image, which we call the semantic embedding of the image. The Network architecture is shown in Fig.~\ref{vgg16}.

\subsubsection{Semantic embedding loss}
We used the L2 norm as a criterion to calculate the semantic embedding differences between the ground truth and the generated image. At the end of the feedforward propagation phase, the semantic embedding loss of the original and generated pictures was extracted by the edited VGG16 network and the L2 loss between the two semantic embeddings will be calculated. The loss obtained after the initial calculation is adjusted by the weight hyperparameter $\lambda_{sEM}$ to restrict the semantic embedding loss to maintain the same order of magnitude as other loss functions to reasonably regulate the impact of this loss on the network update:
\begin{equation}
    \begin{split}
    L_{sEM}=\\
    &\lambda_{sEM}\Sigma^{512}_{i=1}(sEM(X_{GT}) - sEM(X_{pred}))
    \end{split}
\end{equation}\par
where $sEM$ represents the edited VGG semantic embedding extractor, $X_{GT}$ represents the ground truth image,  and $X_{pred}$ represents the generated image.
After that, the scaled semantic embedding losses are added to the overall losses to contribute to the network parameter updates jointly.
\begin{figure}[b!]
\centerline{\includegraphics[width=7.5cm]{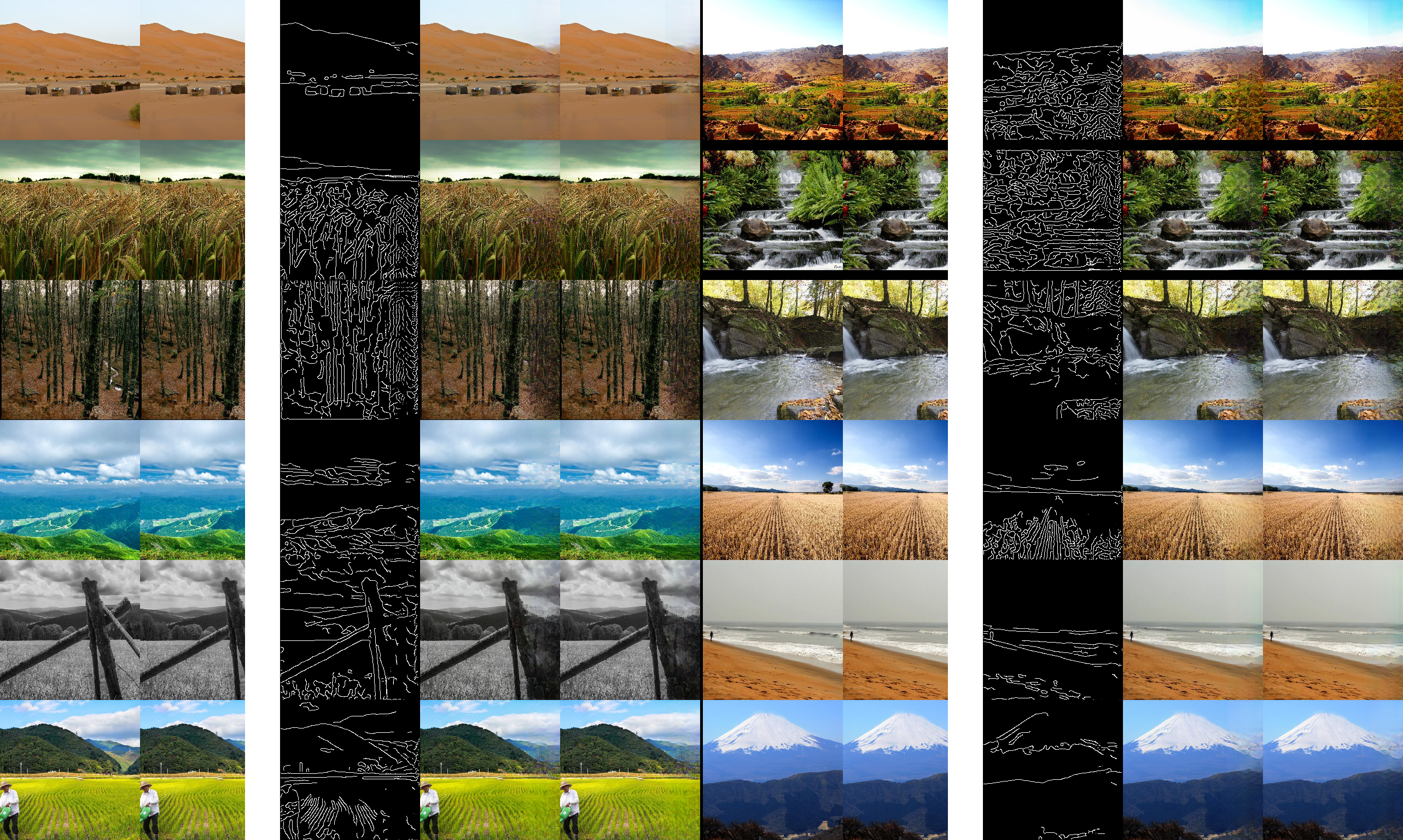}}
\caption{Samples while training.}
\label{samples}
\end{figure}

\begin{figure}[htb!]
\centerline{\includegraphics[width=9cm]{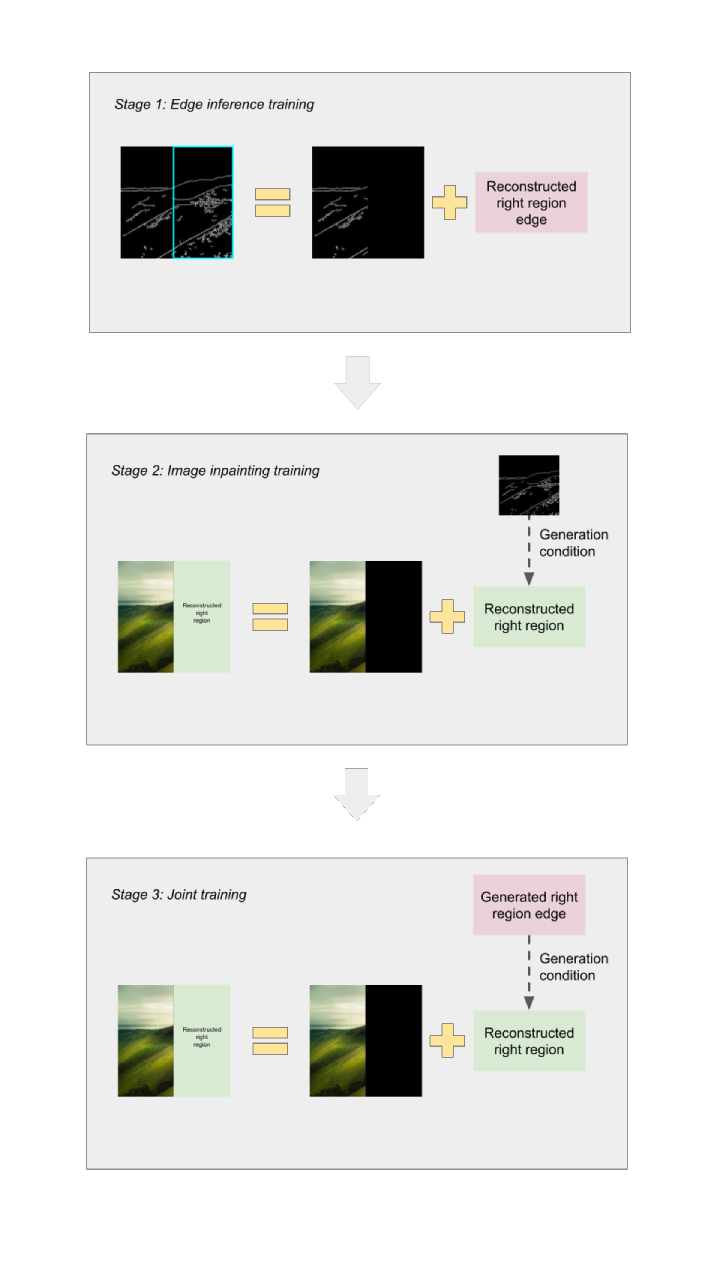}}
\caption{Training pipeline.}
\label{training}
\end{figure}

\subsection{Training}
The holistic training process covers three stages: edge (structure) inference training, contents inpainting training, and joint training. Visualization can be seen as Fig.~\ref{training}. The edge inference training will only handle edge map prediction based on the left legitimate area. The image inpainting training takes the ground truth edge map as a structural prior to generate realistic content. In contrast, the joint training stage takes the edge map generated by the trained edge inference network as a structure prior to outpainting. \par 
The discriminator in every stage determines not only the authenticity of the generated contents but also the degree of matching with the ground truth contents, which will be presented in our design of loss functions. We obtain the reconstructed contents in every stage by combining the right region of generated contents with the left region of ground truth contents.
\begin{equation}
    X_{o}=X_{gt}\bigodot M+X_{pred}\bigodot (1-M)
\end{equation}\par
$X_{o}$ can be outpianting-ed edges or images, while $X_{gt}$ refers to ground truth contents and $X_{pred}$ refers to generated contents. $M$ refers to the binary mask. $\bigodot$ means a Hadamard product.
We sampled some images from training process and show the images in Fig.~\ref{samples}.

\subsubsection{Training configuration}
We fine-tuned a model trained on the Places2 dataset~\cite{zhou2017places} using our landscape dataset~\cite{landscapepictures}. The original pre-trained model has respectively trained 2 million iterations for edge inference and content inpainting, where content inpainting includes training taking ground truth as conditions and training taking generated edge maps as conditions. Additionally, we trained our edge inference model with 75,000 iterations, the content inpainting model taking ground truth edge maps as conditions with 75,000 iterations, and the content inpainting model taking generated edge maps as conditions with 75,000 iterations. All training is done on RTX-2080 Super graphics.

\begin{figure}[b!]
\centerline{\includegraphics[width=7.5cm]{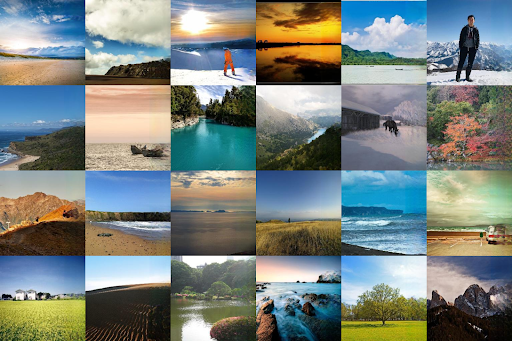}}
\caption{Generation results.}
\label{results}
\end{figure}

\vspace{4mm}
\section{Evaluation}
\subsection{Generation results}
This section will show some generated results from our test set. Randomly sampled generated images are shown in Fig.~\ref{results}. At the present stage, we have only implemented the outpainting of the right quarter of the image on our dataset. The generated results are satisfactory in terms of visual quality. The generated images largely adhere to the structural information of valid parts of the images and are of sensory consistency. Although most images still produce stray colors, artifacts, or unreasonable details, the introduction of semantic embedding loss has alleviated such phenomena to some extent.

\subsection{Effect of semantic embedding loss}
\begin{figure}[t!]
\centerline{\includegraphics[width=9cm]{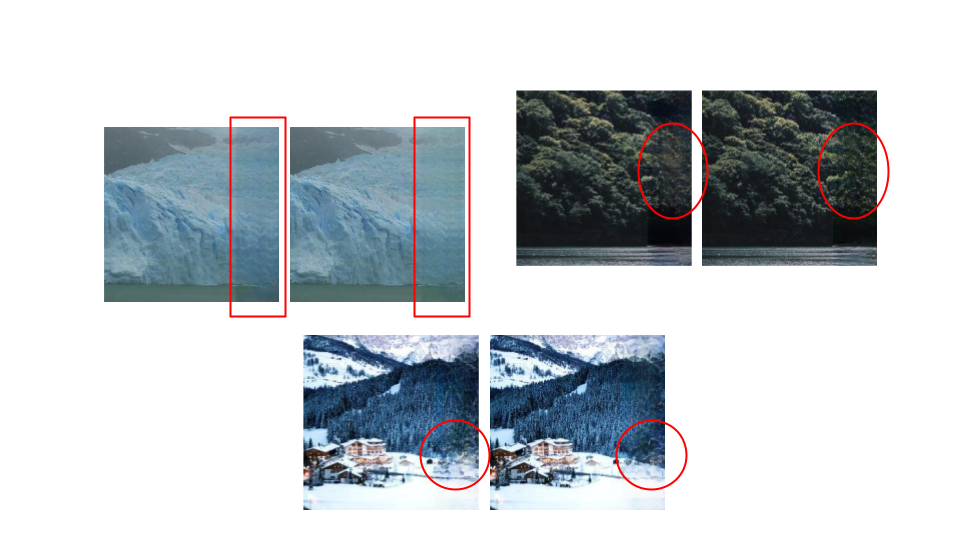}}
\caption{Comparison between plain and improved models.}
\label{comp}
\end{figure}

We add a semantic embedding loss term to motivate the model to focus on the semantic information of the image during the parameter update process so that the semantic information is not overly blurred. The impact of the semantic embedding loss on the model performance is most intuitively seen in the visual effect. With left images representing results from original structure-guided network and right images representing results from semantic-aware model, showing in Fig.~\ref{comp}, we randomly took three generated results from the test set, and we can find that for the left top figure (glacier), the blending of the generated part with the ground truth part appears more seamless, indicating that the colors of the generated images are corrected after the semantic embedding loss is introduced. For the right top figure (forest), we can see that in addition to the color of the generated result being closer to the ground truth, the texture of the image is also more reasonably represented. For the middle lower figure (snowy mountains), the stray colors and artifacts in the labeled areas are reduced and the textures are more closely matched to the original image.\par Quantitatively, we selected three metrics commonly used for image reconstruction quality: Peak Signal-to-Noise Ratio (PSNR), Structural Similarity Index SSIM, and Mean Absolute Error (MAE) to evaluate the generated results. As can be seen from Table.~\ref{quality}, all three metrics improve on the model with the introduction of the new loss term: PSNR improves from 22.702 to 23.127, SSIM improves from 0.890 to 0.894, and MAE decreases from 0.43 to 0.40.

\begin{table}[t]
    \centering
\caption{Image quality evaluation.}
\label{quality}
\begin{tabular}{cccc}
    \toprule
& PSNR & SSIM & MAE \\
    \midrule
Model without $L_{sEM}$ & 22.702 & 0.890 & 0.043\\
Model with $L_{sEM}$ & \textbf{23.127} & \textbf{0.894} & \textbf{0.040}\\

\bottomrule
\end{tabular}
\vspace{-0.3cm}
\end{table}

\vspace{4mm}
\section{Conclusion}
\subsection{Discussion}
For the challenging landscape outpainting task, to compensate for the lack of valid information dimensions compared to the inpainting task, we adopt a state-of-the-art structure-guided generative network that uses the edge maps of the input images as a structural prior to guide content inpainting. The evaluation results show that the structure-guided generative network can indeed generate more realistic and reasonable results. On top of this, we introduce a new loss term called semantic embedding loss, which successfully improves the model performance further in terms of visual quality and quantitative metrics. The remaining problem is that since landscape images generally do not have a meta-structure to accomodate the extraction of structural features, e.g., a distant view with a clear scene segmentation and a close view with a complex texture have different features in the structural dimension most intuitively in the complexity of the edge maps under the same set of parameters. This leads to the inability of the generated images to adapt texture, filling with multi-scale complexity. Similarly, for scene pictures, the semantic content in the pictures is different and varied, and although the current introduction of semantic embedding loss has successfully improved the generation quality, it still does not explicitly preserve the semantic content of the original images.
\subsection{Future works}
We planned three further steps to improve the network. Specifically, the above results showed that the network achieves the outpainting of one-quarter of the image. For the next step, the network will be improved to expand the outpainting to half of the image. Besides, it’s easy to notice that the image with complex edges and colours would result in less quality of the outpainted image (with noise and blur) as mentioned in the discussion section. This is due to the inability to generate accurate edges based on complex edges and the inability to accurately fill colors in the outpainted areas, while the edge generation method will be strongly guided by edge prior to improving the quality of the outpainted image. After proving the effectiveness of multi-stage generation based on structural prior, we plan to further search for more optimized and reasonable structural bootstrap terms. In the future, we plan to experiment with the same process on other structural information (e.g., landmarks or high-frequency components of images) to find a structural term that is most suitable for the landscape picture out-drawing task.






\bibliographystyle{IEEEtran}
\bibliography{main}
%



\end{document}